\documentclass[manuscript]{acmart}
\usepackage{graphicx} 
\usepackage{multirow}
\usepackage{mdframed}
\usepackage{lipsum} % For placeholder text
\usepackage{tabularx}
\usepackage{adjustbox}
\usepackage{booktabs}
\usepackage{hyperref}
\usepackage{tcolorbox}
\AtBeginDocument{%
  }

\setcopyright{none}
\acmConference{}                   % No conference info
\acmYear{}                         % No year
\acmDOI{}                          % No DOI
\acmISBN{}                         % No ISBN

\begin{document}

\title{ConVerSum: A Contrastive Learning-based
Approach for Data-Scarce Solution of
Cross-Lingual Summarization Beyond Direct
Equivalents}

\author{Sanzana Karim Lora}
\email{sanzanalora@yahoo.com}
\author{M. Sohel Rahman}
\email{msrahman@cse.buet.ac.bd}

\author{Rifat Shahriyar}
\email{rifat@cse.buet.ac.bd}
\affiliation{%
  \institution{Bangladesh University of Engineering and Technology}
  \city{Dhaka}
  \country{Bangladesh}
}

\renewcommand{\shortauthors}{Lora et al.}
\renewcommand{\shorttitle}{ConVerSum}

\begin{abstract}
  Cross-lingual summarization (CLS) is a sophisticated branch in Natural Language Processing that demands models to accurately translate and summarize articles from different source languages. Despite the improvement of the subsequent studies, This area still needs data-efficient solutions along with effective training methodologies. To the best of our knowledge, there is no feasible solution for CLS when there is no available high-quality CLS data. In this paper, we propose a novel data-efficient approach, \textbf{ConVerSum}, for CLS leveraging the power of \textbf{con}trastive learning, generating \textbf{ver}satile candidate \textbf{sum}maries in different languages based on the given source document and contrasting these summaries with reference summaries concerning the given documents. After that, we train the model with a contrastive ranking loss. Then, we rigorously evaluate the proposed approach against current methodologies and compare it to powerful Large Language Models (LLMs)- Gemini, GPT 3.5, and GPT 4o proving our model performs better for low-resource languages' CLS. These findings represent a substantial improvement in the area, opening the door to more efficient and accurate cross-lingual summarizing techniques.
\end{abstract}


\iffalse

\fi

\iffalse
\received{20 February 2007}
\received[revised]{12 March 2009}
\received[accepted]{5 June 2009}
\fi
\maketitle

\section{Introduction}
\label{sec:introduction}
\begin{figure}
    \centering
    \includegraphics[width=0.95\linewidth]{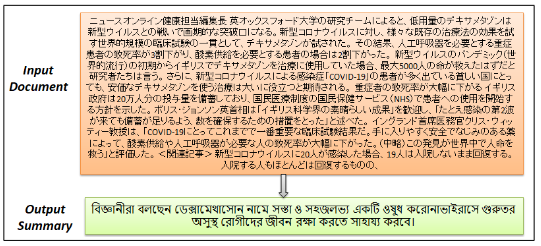}
    \caption{Example of CLS}
    \Description{   }
    \label{fig:CLS}
\end{figure}

Cross-lingual Summarization (CLS) is the process of producing a summary in a different target language (e.g. Bengali) from the source language (e.g. Japanese) (Figure \ref{fig:CLS}). This can aid in easily acquiring the gist of a foreign language article. This task can be thought of as a fusion of machine translation (MT) and monolingual summarization (MS), two open natural language processing (NLP) challenges that have been researched for decades~\cite{PAICE1990171, brown-etal-1993-mathematics}. CLS is an extremely complex process from the perspective of both data and models. In terms of data, as opposed to MS, the lack of naturally occurring documents in a source language and comparable summaries in several target languages make it difficult to gather large-scale and
human-annotated datasets~\cite{ladhak-etal-2020-wikilingua, perez-beltrachini-lapata-2021-models}. From the viewpoint of models, CLS necessitates the ability to translate and summarize, making it challenging to provide precise summaries using CLS independently~\cite{cao-etal-2020-jointly}. Although the absence of high-quality large-scale data has recently been addressed~\cite{bhattacharjee-etal-2023-crosssum}, the CLS still needs models that take data-efficient solutions into account. 

Previous attempts at CLS used pipeline methods such as translate-then-summarize~\cite{Leuski2003_translate_summarize} and summarize-then-translate~\cite{wan-etal-2010-cross}. Due to the necessity for several models, these methods are not only computationally costly but also experience error propagation~\cite{zhu-etal-2019-ncls} among models, which diminishes overall performance rendering it inappropriate for the scenario in real life~\cite{ladhak-etal-2020-wikilingua}. To tackle this problem, CLS has recently been accomplished by utilizing Sequence-to-Sequence(Seq2Seq)~\cite{cho-etal-2014-learning, sutskever2014sequence} and Transformers-based models~\cite{vaswani} to perform the summarization in a single model~\cite{zhu-etal-2019-ncls, Cao_Wan_Yao_Yu_2020, bhattacharjee-etal-2023-crosssum}. The aforementioned achievements have paved the way for the field of CLS research and progressively fueled curiosity in CLS. However, no one proposed any CLS solution when there is no available large-scale CLS corpus to the best of our knowledge. Thus, we wish to investigate the question: \textit{Can we develop a CLS model which will be able to generate coherent and informative summaries when no CLS data is available?}

In this paper, we propose a model \textit{\textbf{ConVerSum}} which provides a data-scarce solution for CLS without parallel corpora by leveraging contrastive learning. At first, we generate versatile candidate summaries in different languages using a seq2seq model, use diverse Beam search to explore word sequences, measure the quality of candidate summaries, apply contrastive learning to minimize positive and maximize negative pairs and train the model with a contrastive ranking loss to improve similarity scores. \textit{ConVerSum} is evaluated using standard metrics and tested on different cross-lingual language pairs and Large Language Models (LLMs). Therefore, our main contribution to this paper is as follows:
\begin{itemize}
    \item We propose a contrastive learning-based approach for CLS \textbf{\textit{ConVerSum}} when there is no large-scale parallel corpus of CLS.
    \item We design a contrastive ranking loss to train the model to improve the similarity score of candidate summaries concerning document and reference summaries.
    \item We carry out a comprehensive comparative study to identify the best-performing approach and prove \textit{ConVerSum}'s robustness on different cross-lingual pairs and low-resource languages. We evaluate it with state-of-the-art LLMs like GPT 3.5, GPT 4o, and Gemini.
\end{itemize}

\section{Related Works}
\label{sec:Related Works}
Over the past five years, more than twenty articles have been published by researchers who did their best to solve the CLS task~\cite{wang2022survey}. Moreover, contrastive learning in abstractive summarization has attracted the attention of researchers lately~\cite{zhang-etal-2022-contrastive-data}. We analyze existing publications to identify the most sophisticated strategies employed in recent studies.  This section provides a full discussion of relevant articles and their working techniques on CLS and contrastive learning in summarization.

\subsection{Pipeline Methods}
\label{subsec:Pipeline Methods}
Early CLS work focused on pipeline approaches, which decompose CLS into MS and MT sub-tasks and complete them sequentially. These approaches can be classified as summarize-then-translate~\cite{orasan-chiorean-2008-evaluation, wan-etal-2010-cross} or translate-then-summarize~\cite{Leuski2003_translate_summarize, wan-2011-using, boudin2011graph, yao-etal-2015-phrase} based on the sequence of completed sub-tasks. The pipeline method is straightforward however has some drawbacks. First of all, they are computationally very expensive for using multiple models; secondly, these techniques are prone to error propagation, resulting in unsatisfactory overall performance~\cite{zhu-etal-2019-ncls}. Moreover, latency occurs during inference. Additionally, training MT models requires either a big corpus or commercial services.

\subsection{End-to-End Methods}
\label{sec:End-to-End Methods}
Advancements in seq2seq~\cite{cho-etal-2014-learning, sutskever2014sequence} and Transformer-based models~\cite{vaswani} have led to the invention of end-to-end methods for performing CLS with a single model to resolve the drawbacks mentioned in Section \ref{subsec:Pipeline Methods}. The end-to-end methods can be classified into four categories - multi-task framework, knowledge distillation framework, resource-enhanced framework and pre-training framework.

Multi-task is a machine learning approach in which a single model is designed by combining several related tasks. Several studies train unified models using CLS in conjunction with related tasks such as MT and MS~\cite{zhu-etal-2019-ncls, cao-etal-2020-jointly, Takase2020MultiTaskLF, bai-etal-2021-cross, liang-etal-2022-variational}. CLS models thus gain from the related tasks as well.

Knowledge distillation is a machine learning approach used to reduce a large, complicated model to a smaller, simpler one. This is accomplished by transferring information from the larger model, known as the teacher model, to the smaller model, known as the student model~\cite{hinton2015distilling}. To teach the CLS model in the knowledge-distillation framework, several researchers employ either MT/MS or both models because to have a mutual connection. In addition to CLS labels, the student model can gain knowledge from the output or hidden state of the teacher's models. Extensive MS and MT datasets are leveraged to train MS and MT models, respectively which are bidirectional GRU~\cite{cho-etal-2014-learning}. These models act as teachers for a CLS student model, with KL-divergence aligning generation probabilities and maximizing the training objective~\cite{ayana2018}. Afterward, transformer~\cite{vaswani} is used as the backbone for the CLS student model and the MS teacher model and further trains the student model~\cite{duan-etal-2019-zero}.

The resource-enhanced framework enriches input documents with extra resources. Output summaries are generated based on encoded and enriched information. Translation patterns in CLS are investigated by encoding source language texts using a transformer encoder~\cite{zhu-etal-2020-attend}. The TextRank toolset~\cite{mihalcea-tarau-2004-textrank} helps create article graphs by extracting important information from input sequences~\cite{jiang2022cluegraphsum}. A transformer-based clue encoder and a graph neural network-based graph encoder are used, respectively, to encode these clues and article graphs. Ultimately, the translation distribution from~\cite{zhu-etal-2020-attend} is included in final summaries generated by a transformer decoder with dual cross-attention methods.

In the context of NLP, pre-trained models are neural network architectures that are extensively trained on textual data to identify linguistic patterns and semantic representations. These models are trained on extensive text corpora, which helps them understand word and phrase semantic meanings as well as contextual relationships. By using pre-trained NLP models, transfer learning can be accomplished more successfully. This is because the gained information can be fine-tuned on certain NLP tasks improving efficiency and performance. The pre-trained models revolutionize NLP~\cite{Qiu_2020}.

Recently, multi-lingual pre-trained generative models have performed exceptionally well in various multi-lingual NLP tasks. mBART~\cite{liu-etal-2020-multilingual-denoising} is a multilingual pre-trained model which is another variant of BART~\cite{lewis-etal-2020-bart} model developed by Facebook AI. mBART is pre-trained with denoising objectives based on the BART architecture and vast unlabeled multilingual datasets. Specifically, it is trained in over 25 languages text data. Initially excelling in MT~\cite{liu-etal-2020-multilingual-denoising}, a recent study shows that it can outperform many multi-task CLS models on large-scale datasets with straightforward fine-tuning~\cite{liang-etal-2022-variational}. Subsequently, the language processing capabilities of mBART are expanded from 25 to 50 languages with mBART-50~\cite{tang-etal-2021-multilingual}. Aside from the BART-style pre-trained models, mT5~\cite{xue-etal-2021-mt5} shines out as a multilingual version of T5~\cite{t5_raffel2020}, trained across 101 languages with the T5-style span corruption objective. Despite the remarkable performance of these general pre-trained models, they lack explicit cross-lingual supervision and instead rely only on denoising or span corruption objectives across several languages. They have thus far not fully explored their cross-lingual potential.

\subsection{Contrastive Learning for Monolingual Abstractive Summarization}
Contrastive learning has recently successfully emerged in various fields of NLP~\cite{zhang-etal-2022-contrastive-data}, to learn good and discriminative representations. Contrastive learning is a self-supervised approach to training models to differentiate between samples that are similar versus dissimilar. This approach is popularized in the SimCLR framework~\cite{chen2020simple}, which uses contrastive loss to minimize the distance between representations for similar data points and maximize it for dissimilar points.

The application of contrastive learning for abstractive summarization is a nascent yet promising branch of research. The SimCLS framework is designed to bridge the gap between standard seq2seq models and learning objectives by utilizing contrastive learning to adapt to evaluation metrics with the learning objective~\cite{liu-liu-2021-simcls}. A contrastive attention mechanism is proposed for abstractive sentence summarization that builds on the seq2seq framework by combining conventional attention, which focuses on relevant parts of the source sentence, and opponent attention, which focuses on less relevant parts, with both trained in opposite directions using novel softmax and softmin functions~\cite{duan-etal-2019-contrastive}. COLO, a contrastive learning-based re-ranking framework for one-stage summarizing is introduced that aligns training and assessment by producing summaries based on summary-level scores without the requirement for extra modules or parameters~\cite{an-etal-2022-colo}. CALMS, a unified summarizing model is constructed which is trained with contrastive learning by using MLGSum, a large-scale multilingual summarizing corpus including 1.1 million articles and summaries in 12 languages~\cite{wang2021contrastive}. It incorporates contrastive sentence ranking and sentence-aligned replacement to improve extractive skills and align sentence-level representations across languages.

\section{ConVerSum}
\label{sec:conversum}
\begin{figure}
    \centering
    \fbox{\includegraphics[width=0.95\linewidth]{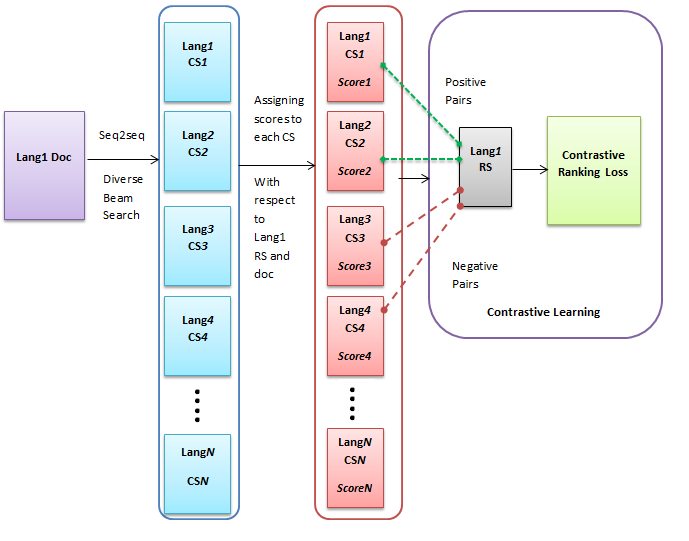}}
    \caption{General Structure of \textit{ConVerSum}. Here CS, RS, Lang and Doc refer to candidate summary, reference summary, language and document, respectively. For better realization, we assume score1 > score2 > score3 > score4 \ldots scoreN.}
    \Description{   }
    \label{fig:general_structure_of_proposed_methoodology}
\end{figure}

In this section, we elaborate on our proposed approach \textit{ConVerSum} which is categorized into several phases. Figure \ref{fig:general_structure_of_proposed_methoodology} illustrates the general structure of \textit{ConVerSum}. \textit{ConVerSum} deals with the data scarcity
challenges where the existing models require high-quality large-scale datasets as discussed in
Section \ref{sec:Related Works}. 

\subsection{First Phase: Candidate Summary Generation in Different Languages} 
\label{sec:First Phase:Candidate Summary Generation in Different Languages}
In this phase, candidate summaries are generated in different languages using monolingual languages. We generate candidate summaries inspired by the candidate generation process of SimCLS~\cite{liu-liu-2021-simcls}. Though SimCLS is developed for monolingual abstractive summarization, their candidate generation process motivates us to create a robust CLS model. The detailed procedure is as follows.

\subsubsection{Seq2Seq Model Selection}
\label{subsec:Seq2Seq_Model_Selection}
A renowned seq2seq model, mT5~\cite{xue-etal-2021-mt5} is utilized to generate multiple candidate summaries in different languages from the given monolingual document. mT5 has multilingual capabilities as it is pre-trained on a vast and diversified multilingual dataset spanning more than 100 languages. This thorough training enables the model to create summaries in many languages from a single monolingual source, making it very versatile and adaptable to a variety of linguistic settings. The enhanced many-to-many(m2m) mT5 checkpoint\footnote{\url{csebuetnlp/mT5_m2m_crossSum_enhanced}} fine-tuned on all cross-lingual pairs of CrossSum~\cite{bhattacharjee-etal-2023-crosssum} dataset is employed in this case.

\subsubsection{Diverse Summary Exploration}
Many candidate summaries can be generated by utilizing the seq2seq model as described in Section \ref{subsec:Seq2Seq_Model_Selection}. However, not all generated candidate summaries are high-quality. Moreover, there is a possibility of generating repetitive summaries and all summaries in the same language. To tackle these issues, diverse Beam search~\cite{diverse_beam_search} is used during the generation of candidate summaries which explores different sequences of words in the generated summary, contributes to generating diverse candidate summaries, and finds out the high-quality summaries.   

\subsection{Second Phase: Candidate Summary Quality Measurement}
In this phase, the generated candidate summaries as described in Section \ref{sec:First Phase:Candidate Summary Generation in Different Languages} are measured concerning the reference summaries and the given documents. The detailed process is as follows:

\subsubsection{Evaluation Function Initialization}
The evaluation function is initiated using XLM-RoBERTa~\cite{conneau-etal-2020-unsupervised-xlm}. The base model of XLM-RoBERTa\footnote{\url{FacebookAI/xlm-roberta-base}} available in huggingface transformers library\footnote{\url{https://huggingface.co/}} is specifically employed in this case. The purpose of utilizing XLM-RoBERTa is to map semantically similar text from various languages into a shared embedding space. Encoding the candidate summaries, reference summaries, and documents can capture the meaning and context of the text. 

\subsubsection{Similarity Calculation}
For similarity calculation, cosine similarity is utilized~\cite{rahutomo2012semantic}. The purpose of calculating similarity is to identify how similar the candidate summaries and the reference summary are to the given source document. The general representation of the similarities between the source document, reference summary, and candidate summaries is as follows where \textit{C}, \textit{R} and \textit{S} represent the candidate summaries, reference summary, and the source document, respectively.
\[
\text{Similarity} = f(C, R, S) = \frac{C \cdot S + R \cdot S}{\|C\| \|S\| + \|R\| \|S\|}
\]
This score is utilized to measure the contrastive ranking loss described in Section \ref{sec:Contrastive Ranking Loss}. 

\subsubsection{Candidate Summary Scoring}
While generating candidate summaries as described in Section \ref{sec:First Phase:Candidate Summary Generation in Different Languages}, LaSE~\cite{bhattacharjee-etal-2023-crosssum}, which is specifically designed for evaluating CLS is computed concerning the reference summary. That means a LaSE score is assigned to each candidate summary and sorted in descending order. This indicates the highest-scoring candidate is more similar to the reference summary and the lowest-scoring candidate is less similar to the reference summary. This process can be defined as follows where $C_i$ and \textit{R} represent the \textit{i-th} candidate summary and reference summary, respectively.
\[
\text{LaSE}(S_1) \geq \text{LaSE}(S_2) \geq \dots \geq \text{LaSE}(S_n) = f(C_i, R) \quad \text{for each} \ i \in \{1, 2, \dots, n\}
\]

Both cosine similarity and LaSE scores are used in the evaluation of candidate summaries to capture various characteristics of summary quality. Every metric offers distinct perspectives that enhance one another, resulting in a more thorough assessment as there is a possibility of generating bias using only one scoring method as CLS is an extremely complex task. Striking a balance between semantic coherence and exact content overlap enables a thorough examination and guarantees that the summaries that are produced are correct and relevant. This thorough method produces summaries that are of higher quality and more dependability, which is important for tasks requiring accurate factual information in addition to contextual knowledge.

\subsection{Third Phase: Contrastive Learning Approach}
We apply a contrastive learning objective that cultivates a common representation space to improve the quality of cross-lingual summary generation. The objective of this approach is to maximize the distance between negative pairings and minimize the distance between positive pairs. The model is trained to push negative pairings farther apart and bring positive pairs closer together in the representation space by employing a contrastive ranking loss function. This ensures that the generated summaries are both semantically correct and pertinent to the given context. The model's capacity to differentiate between and provide excellent summaries is much enhanced by this approach. In this section, the whole process is described elaborately.

\subsubsection{Positive-Negative Pair Construction}
To implement the contrastive learning objective, we need to construct the pairs that are used for positive and negative samples for contrastive learning. To properly evaluate the quality of candidate summaries, we use a positive-negative pair creation procedure. This entails computing the scores of candidate summaries for the provided documents. Positive pairings include candidate summaries that are more comparable to the reference summary, implying greater quality. In contrast, negative pairings comprise summaries that are of inferior quality when compared to other candidate summaries and the reference summary. This strategy aids in discriminating between better and inferior summaries, hence improving the model's capacity to produce high-quality summaries.

\subsubsection{Contrastive Ranking Loss}
\label{sec:Contrastive Ranking Loss}
The goal of the Contrastive Ranking Loss approach is to maximize the correlations between rankings of various candidate summaries. This is accomplished by adding a margin, which acts as a threshold to control the amount of space between summaries in pairs. This margin makes sure that summaries that should be similar are represented by positive pairings, which are encouraged to have high similarity scores, while summaries that should be dissimilar are represented by negative pairs, which are encouraged to have low similarity scores. The approach efficiently separates pertinent and extraneous summaries by enforcing this margin. Furthermore, Contrastive Ranking Loss promotes proper ranking connections by punishing deviations from the intended ranking order. This means that summaries should be ordered based on their relevance or similarity to an expected reference point. By optimizing the loss function using these ranking connections, the technique guarantees that the most relevant summaries are prioritized in the output. This loss can be mathematically represented as follows:
\[
\mathcal{L} = \sum_{i=1}^{n} \max(0, \text{margin} - f(C_i, R^+) + f(C_i, R^-))
\]
where, $R^+$ and $R^-$ represent the positive and negative pairs, respectively.

\subsection{Fourth Phase: Model Training}
The model is trained on the cross-lingual summary generation task by using the contrastive ranking loss described in Section \ref{sec:Contrastive Ranking Loss}. This process entails modifying the model's parameters to successfully capture the complex connections between the source documents and their related summaries across multiple languages. This is accomplished by using Contrastive Loss as the optimization goal, which provides a margin to encourage the model to differentiate between important and irrelevant summary pairs. During training, the model learns to give high similarity scores to positive summary pairings that transmit comparable information across languages and low similarity scores to negative pairs that include different documents.

\subsection{Fifth Phase: Fine-Tuning and Evaluation}
By fine-tuning the model, it obtains the capacity to create short and informative summaries that keep the substance of the original material across language limitations, thereby increasing the performance of cross-lingual summarizing tasks. Then, the model's effectiveness is evaluated by the standard evaluation metrics LaSE~\cite{bhattacharjee-etal-2023-crosssum} and BERTScore~\cite{bert-score}.

\subsection{Sixth Phase: Model Verification}
As we developed our model on a monolingual high-resource dataset, we need to verify the model's robustness on the existing CLS dataset. Furthermore, we explore our approach to different low-resource monolingual datasets to ensure adaptability in scenarios when there is no parallel CLS data. Moreover, we make a performance comparison of our model to the most powerful latest LLMs.

Furthermore, we repeat the whole process using Flan-T5~\cite{flan_t5} instead of mT5 where we use the checkpoint\footnote{\url{google/flan-t5-large}} to identify which approach performs better. All the details are explained in Section \ref{sec:Experimental Evaluation}.

\section{Experimental Evaluation}
\label{sec:Experimental Evaluation}

\subsection{Training Setup}
The main computing resource for model training is the NVIDIA GeForce RTX 3090 GPU. We use the broadly recognized deep learning framework PyTorch\footnote{\url{https://pytorch.org/}} (version 3.7.9) to develop \textit{ConVerSum}.

\subsection{Training Consideration}
\label{subsec:Training Consideration}
We have two things in our consideration for training \textit{ConVerSum}. The first is to train \textit{ConVerSum} utilizing a large-scale monolingual dataset of high-resource languages. The second is to train \textit{ConVerSum} utilizing a significantly smaller monolingual dataset of low-resource languages. The reason behind considering these two options is to demonstrate how \textit{ConVerSum} works while training the model on a large-scale dataset of high-resource language and a significantly smaller dataset of low-resource language which proves the power of contrastive learning for both large-scale and small-scale datasets. For both cases, we analyze the performance of \textit{ConVerSum} on a cross-lingual dataset to verify the model's robustness.

\subsection{Datasets}
\label{subsec:datasets}
According to the consideration mentioned in Section \ref{subsec:Training Consideration}. We choose CNN/DailyMail~\cite{hermann2015teaching, nallapati-etal-2016-abstractive} for the high-resource large-scale monolingual dataset and XLSUM~\cite{hasan-etal-2021-xl} for low-resource significantly smaller monolingual datasets. We choose the monolingual Bengali, Thai, Burmese, and Tigrinya language datasets for our task from XLSUM. Finally, we perform a performance analysis of our model on CrossSum~\cite{bhattacharjee-etal-2023-crosssum}, a large-scale  CLS dataset comprising 1.68
million article-summary samples in 1,500+ language pairs.
The dataset for training and verification of the model is shown in Table \ref{tab:training-verification-datasets}. The dataset statistics for monolingual datasets for training are shown in Table \ref{tab:monolingual-datasets-distribution}. The dataset splits are publicly
available and provided by the author and intended for efficient training, validation, and unbiased
testing to ensure robust model performance across various languages and summarization tasks.

\begin{table}[ht]
\centering
\caption{Datasets for Training and Verification}
\label{tab:training-verification-datasets}
\begin{tabular}{>{\raggedright\arraybackslash}p{4cm} >{\raggedright\arraybackslash}p{4cm}}
\toprule
\textbf{Training Dataset (Monolingual)} & \textbf{Verification Dataset (Cross-lingual)} \\
\midrule
CNN/DailyMail & \\
XLSUM & CrossSum \\
\hspace{1em}• Bengali & \\
\hspace{1em}• Thai & \\
\hspace{1em}• Burmese & \\
\hspace{1em}• Tigrinya & \\
\bottomrule
\end{tabular}
\end{table}

\begin{table}[ht]
\centering
\caption{Training, Validation, and Test Splits for Monolingual Datasets for Training}
\label{tab:monolingual-datasets-distribution}
\begin{tabular}{rlrrr}
\toprule
\textbf{Resource-Dataset} & \textbf{Monolingual Data} & \textbf{Train} & \textbf{Val} & \textbf{Test} \\
\midrule
high-large & CNN/DailyMail & 287113 & 13368 & 11490 \\
low-small & XLSUM-Bengali & 8102 & 1012 & 1012 \\
low-small & XLSUM-Thai & 6616 & 826 & 826 \\
low-small & XLSUM-Burmese & 4569 & 570 & 570 \\
low-small & XLSUM-Tigrinya & 5451 & 681 & 681 \\
\bottomrule
\end{tabular}
\end{table}

\subsection{Baseline Model}
\label{subsec:baseline}
We use the mT5~\cite{xue-etal-2021-mt5} and FLAN-T5~\cite{flan_t5} models as baselines for our tasks. These models are selected for their effectively handled multilingual data and demonstrated state-
of-the-art performance in tasks related to NLP, such as text summarization. 

\subsection{Evaluation Metrics}
We use LaSE~\cite{bhattacharjee-etal-2023-crosssum} and BERTScore~\cite{bert-score}. LaSE is specifically designed for evaluating CLS considering meaning similarity, language confidence and length penalty. BERTScore exploits the pre-trained contextual embeddings from BERT~\cite{devlin-etal-2019-bert} to identify words in candidate and reference sentences based on cosine similarity.

\subsection{First Consideration: Training \textit{ConVerSum} on High-Resource Large
Scale Dataset}
\label{subsec:Training on High-Resource Large
Scale Dataset}
This section elaborates on how we develop \textit{ConVerSum} on CNN/DailyMail dataset
as we discussed in Section \ref{subsec:Training Consideration} and \ref{subsec:datasets}. We utilize mT5 and Flan-T5 separately as they are our baselines for developing \textit{ConVerSum}. Then, we perform a
comparison analysis to know which Transformer assists us in developing the robust model.

To generate candidate summaries in different languages, we use mT5 checkpoint\footnote{\url{https://huggingface.co/csebuetnlp/mT5_m2m_crossSum_enhanced}} available in Huggingface\footnote{\url{https://huggingface.co/}}. For diversity sampling using diverse Beam search, we employ 8 groups, yielding 8 candidate summaries. The maximum length of the generated candidate summaries is specified at 80 words. Using an identical length guarantees uniformity between summaries in different languages. This uniformity makes the evaluation process easier and allows for fair comparisons. Moreover, languages differ in terms of word length and sentence structure. The summary word restriction guarantees all summaries are reasonably equal in terms of their content and depth, independent of language. For validation and test sets, we also generate the candidate summaries. To encode reference summaries, candidate summaries, and documents, we initiate the evaluation function with the XLM-RoBERTa checkpoint\footnote{\url{FacebookAI/xlm-roberta-base}} as it learns the inner representation of 100 languages. Then, we apply the contrastive learning objective described in Section \ref{sec:conversum}. For training, Adam~\cite{adam_2015} optimizer is used with learning rate scheduling. The batch size is set to 2 for candidate generation and 4 for model training after trying 32, 16, 8 and 4 because of our GPU limitations. This indicates that we always choose the maximum batch size allowed by our GPU memory. We evaluate the performance of \textit{ConVerSum} on the validation set at every 1000 steps using the LaSE. We set 15 epochs for training.

For experimenting with Flan-T5, we use Flan-T5 checkpoint\footnote{\url{https://huggingface.co/google/flan-t5-large}} to generate candidate summaries in different languages. The other settings like the dataset used for training, and the overall training process with all setups remain as same as described before. This consistent setup makes it easier to do comprehensive research and comparisons across the models of CLS.

\subsection{Experimental Results on High-Resource Large Scale Dataset}
\label{subsec:Experimental Results on High-Resource Large Scale Dataset}
This section conveys the experimental findings from the analysis of the \textit{ConVerSum}s along with baselines for the high-resource large-scale dataset. The models' performance is evaluated using standard metrics like LaSE and BERTScore, which serve as a complete assessment of the quality of the generated summaries. There is no significant difference between the training time for the baseline and \textit{ConVerSum}. The following sections describe the experimental results, providing insights into the proposed methodologies' effectiveness and reliability for high-resource large-scale dataset.

\subsubsection{Model Performance on CNN/DailyMail Test Set}
The performance of \textit{ConVerSum} based on both mT5 and Flan-T5 on the test set of the CNN/DailyMail dataset is shown in Table \ref{tab:mt5_on_cnndm} and Table \ref{tab:flant5_on_cnndm}, respectively. 

\begin{table}[]
\centering
\caption{Performance of baseline and \textit{ConVerSum} based on mT5 on the Test set. When we generate candidate summaries using mT5, we choose mT5 as the baseline model to compare our model with it. LaSE, designed for specifically CLS evaluation, illustrates the model's robustness on CLS, which is the best choice for CLS evaluation, and BERTScore illustrates the semantic similarity used as a supplementary metric. }
\label{tab:mt5_on_cnndm}
\begin{tabular}{ccc}
\toprule
\textbf{\begin{tabular}[c]{@{}c@{}}Evaluation \\ Score\end{tabular}} & \textbf{Baseline (mT5)} & \textbf{\begin{tabular}[c]{@{}c@{}}ConVerSum\\ on mT5\end{tabular}} \\
\midrule
\textbf{LaSE}                                                            & 0.3886                  & \textbf{0.4408}                                                          \\
\textbf{BERTScore }                                                           & 0.8315                  & \textbf{0.8363}       \\
\bottomrule
\end{tabular}
\end{table}

\begin{table}[]
\centering
\caption{Performance of baseline and \textit{ConVerSum} based on Flan-T5 on the Test set. When we generate candidate summaries using Flan-T5, we choose Flan-T5 as the baseline model to compare our model with it. LaSE, designed for specifically CLS evaluation, illustrates the model's robustness on CLS, which is the best choice for CLS evaluation, and BERTScore illustrates the semantic similarity used as a supplementary metric.}
\label{tab:flant5_on_cnndm}
\begin{tabular}{ccc}
\toprule
\textbf{\begin{tabular}[c]{@{}c@{}}Evaluation \\ Score\end{tabular}} &
  \textbf{Baseline (Flan T5)} &
  \textbf{\begin{tabular}[c]{@{}c@{}}ConVerSum\\ on Flan T5\end{tabular}} \\
\midrule
\textbf{LaSE} &
  0.3585 &
  \textbf{0.4012} \\
\textbf{BERTScore} &
  0.8411 &
  0.8381
\\
\bottomrule
\end{tabular}
\end{table}

In comparison to the mT5 and Flan-T5 baselines, \textit{ConVerSum} performs better, as demonstrated by the findings shown in Table \ref{tab:mt5_on_cnndm} and Table \ref{tab:flant5_on_cnndm}. The \textit{ConVerSum}'s BERTScore on Flan-T5 is almost identical to the baseline's, however, the overall performance metrics show a distinct benefit. Furthermore, \textit{ConVerSum} regularly performs better than the Flan-T5  when compared to the mT5.

\subsubsection{Performance Comparison of \textit{ConVerSum} with Baselines on CrossSum Dataset}
\label{subsubsec:Performance Comparison of ConVerSum with Baselines on Cross-Lingual Dataset}

To evaluate the performance of \textit{ConVerSum} and compare it with the baselines mentioned in Section \ref{subsec:baseline}, we consider three types of combinations - \textit{(i) Source English to Target Other Languages, (ii) Source Other Languages to Target English and (iii) Source Other Languages to Target Other Languages Except English} and observe the performance of all of these combinations. Table \ref{tab:m5_on_crosssum_en_others}, Table \ref{tab:ft5_on_crosssum_en_others}, Table \ref{tab:m5_on_crosssum_others_en}, Table \ref{tab:ft5_on_crosssum_others_en}, Table \ref{tab:m5_on_crosssum_others_others} and Table \ref{tab:ft5_on_crosssum_others_others} demonstrate the performance comparison for these combinations.

\begin{table}[]
\centering
\caption{Performance comparison of (English source and other target) baseline and \textit{ConVerSum} based on mT5 on CrossSum. When we generate candidate summaries using mT5, we choose mT5 as the baseline model
to compare our model with it. LaSE, designed for specifically CLS evaluation, illustrates the
model’s robustness on CLS, which is the best choice for CLS evaluation, and BERTScore
illustrates the semantic similarity used as a supplementary metric.}
\label{tab:m5_on_crosssum_en_others}
\begin{tabular}{lllll}
\toprule 
                        & \multicolumn{2}{l}{\textbf{Baseline (mT5)}} & \multicolumn{2}{l}{\textbf{ConVerSum}} \\
\multirow{-2}{*}{\textbf{\begin{tabular}[c]{@{}l@{}}Source-target    \\ (English - others)\end{tabular}}} &
  \textbf{LaSE} &
  \textbf{BERTScore} &
  \textbf{LaSE} &
  \textbf{BERTScore} \\
  \midrule
English\ - Chinese\_simplified & 0.5638           & 0.8573          & \textbf{0.5987 }          & 0.8556          \\

English\ - Hindi               & 0.6042                    & 0.8510          & \textbf{0.6398 }          & \textbf{0.8567 }         \\
English\ - Bengali                  & 0.5874                    & 0.8407          & \textbf{0.6083 }          & \textbf{0.8440 }         \\
 
English\ - Japanese            & 0.5773                    & 0.8364          & \textbf{0.5879 }          & \textbf{0.8384 }         \\
English\ - Marathi             & 0.6071                    & 0.8433          & \textbf{0.6305 }          & \textbf{0.8518 }         \\

English\ - Thai                & 0.4121                    & 0.8324          & \textbf{0.4606 }          & 0.8323          \\
English\ - Tigrinya            & 0.5532                    & 0.8062          & \textbf{0.5740}           & 0.8056       \\
\bottomrule
\end{tabular}
\caption*{\textit{\textbf{Key Findings: }ConVerSum outperforms the baseline for LaSE, ensuring CLS robustness. Besides, ConVerSum performs better/almost similar to the baseline for BERTScore emphasizing the meaning similarity.} }
\end{table}

%%%%%% flan T5 table starts %%%%%%%%%

\begin{table}[]
\centering
\caption{Performance comparison of (English source and Others target) baseline and \textit{ConVerSum} based on Flan-T5 on CrossSum. When we generate candidate summaries using Flan-T5, we choose Flan-T5 as the baseline model
to compare our model with it. LaSE, designed for specifically CLS evaluation, illustrates the
model’s robustness on CLS, which is the best choice for CLS evaluation, and BERTScore
illustrates the semantic similarity used as a supplementary metric.}
\label{tab:ft5_on_crosssum_en_others}
\begin{tabular}{lllll}
\toprule
                     & \multicolumn{2}{l}{\textbf{Baseline (Flan-T5)}} & \multicolumn{2}{l}{\textbf{ConVerSum}} \\
\multirow{-2}{*}{\textbf{\begin{tabular}[c]{@{}l@{}}Source-target\\    (English   to   others)\end{tabular}}} &
  \textbf{LaSE} &
  \textbf{BERTScore} &
  \textbf{LaSE} &
  \textbf{BERTScore} \\
  \midrule
English - Chinese\_simplified &  0.3814 &
  0.8130 &
  \textbf{0.3890} &
  0.8097 \\

English -   Hindi    & 0.3700                 & 0.8124                 & \textbf{0.3707}           & 0.8063          \\
English -   Bengali  & 0.3341                 & 0.7967                 & 0.3328           & \textbf{0.8015}          \\

English - Japanese   & 0.3464                 & 0.7995                 & \textbf{0.3570 }          & 0.7960          \\
English - Marathi    & 0.3870                 & 0.8046                & 0.3750           & 0.7988          \\

English -   Thai     & 0.2971                 & 0.7944                & \textbf{0.3019 }          & 0.7907          \\
English -   Tigrinya & 0.4352                 & 0.7773                & 0.3193           & 0.7722      \\
\bottomrule
\end{tabular}
\caption*{\textit{\textbf{Key Findings: }ConVerSum performs better/almost similar to the baseline for LaSE. Besides, \textit{ConVerSum} performs almost similarly to the baseline for BERTScore. However, it has significantly lower performance than \textit{ConVerSum} on mT5(mentioned in Table \mbox{\ref{tab:m5_on_crosssum_en_others}}) as the generated candidates fail to show diverse quality and languages for Flan-T5.} }
\end{table}

%%%%% flant5 table ends %%%%

%%%%others to en
\begin{table}[]
\centering
\caption{Performance comparison of (Others source and English target) baseline and \textit{ConVerSum} based on mT5 on CrossSum. When we generate candidate summaries using mT5, we choose mT5 as the baseline model to compare our model with it. LaSE, designed for specifically CLS evaluation, illustrates the
model’s robustness on CLS, which is the best choice for CLS evaluation, and BERTScore illustrates the semantic similarity used as a supplementary metric.}
\label{tab:m5_on_crosssum_others_en}
\begin{tabular}{lllll}
\toprule
                              & \multicolumn{2}{l}{\textbf{Baseline (mT5)}} & \multicolumn{2}{l}{\textbf{ConVerSum}} \\
\multirow{-2}{*}{\textbf{\begin{tabular}[c]{@{}l@{}}Source-target\\    (others to English)\end{tabular}}} &
  \textbf{LaSE} &
  \textbf{BERTScore} &
  \textbf{LaSE} &
  \textbf{BERTScore} \\
  \midrule
Chinese\_simplified - English & 0.4742          & 0.8418          & \textbf{0.5361}           & \textbf{0.8437}          \\

Hindi - English               & 0.5179                    & 0.8335          & \textbf{0.5239}           & 0.8329          \\
Bengali   - English           & 0.5168                    & 0.8263         & 0.5149           & 0.8258          \\

Japanese - English            & 0.5216                    & 0.8341          & \textbf{0.5548}           & \textbf{0.8347 }         \\
Marathi - English             & 0.5177                    & 0.8290          & \textbf{0.5341}           & 0.8275          \\

Thai - English                & 0.5078                    & 0.8298          & \textbf{0.5223}           & \textbf{0.8315}          \\
Tigrinya - English            & 0.4478                   & 0.7935          & 0.3758           & \textbf{0.7936}       \\
\bottomrule
\end{tabular}
\caption*{\textit{\textbf{Key Findings: }ConVerSum shows significantly improved performance for Chinese\_simplified-English. However, it shows lower than baseline performance for Tigrinya-English. Moreover, our model shows almost similar to baseline performance for Bengali-English as the generated candidates fail to show diverse quality summaries in Tigrinya and Bengali. Besides, \textit{ConVerSum} outperforms the baseline for the rest of the pairs.} }
\end{table}

%%%%%%%%%flan t5 tables starts %%%%%%%%%%

\begin{table}[]
\centering
\caption{Performance comparison of (Others source and English target) baseline and \textit{ConVerSum} based on Flan-T5 on CrossSum. When we generate candidate summaries using Flan-T5, we choose Flan-T5 as the baseline model
to compare our model with it. LaSE, designed specifically for CLS evaluation, illustrates the
model’s robustness on CLS, which is the best choice for CLS evaluation, and BERTScore
illustrates the semantic similarity used as a supplementary metric.}
\label{tab:ft5_on_crosssum_others_en}
\begin{tabular}{lllll}
\toprule
                                & \multicolumn{2}{l}{\textbf{Baseline (Flan-T5)}} & \multicolumn{2}{l}{\textbf{ConVerSum}} \\
\multirow{-2}{*}{\textbf{\begin{tabular}[c]{@{}l@{}}Source-target\\    (others to English)\end{tabular}}} &
\textbf{LaSE} &
\textbf{BERTScore} &
\textbf{LaSE} &
\textbf{BERTScore} \\
\midrule
Chinese\_simplified   - English & 0.1181                 & 0.6635                 & \textbf{0.1425}           & 0.5891          \\

Hindi - English                 & 0.1332                 & 0.6760                & \textbf{0.1352}           & 0.5537          \\
Bengali   -   English           & 0.1292                & 0.6879                 & 0.1187           & 0.5593          \\
 
Japanese - English              & 0.2168                 & 0.5592                 & \textbf{0.2442}           & 0.3927          \\
Marathi - English               & 0.1278                 & 0.6678               & \textbf{0.1315}           & 0.5815          \\

Thai - English                  & 0.1359                 & 0.6809                 & \textbf{0.1748 }          & 0.6603          \\
Tigrinya - English              & 0.1039                 & 0.5217                 & \textbf{0.1088 }          & 0.3877 \\
\bottomrule
\end{tabular}
\caption*{\textit{\textbf{Key Findings: }ConVerSum outperforms baseline for LaSE except Bengali-English
However, it shows significantly lower performance than \textit{ConVerSum} on mT5 (mentioned in Table \mbox{\ref{tab:m5_on_crosssum_others_en}}) as the generated candidates fail to show diverse quality and languages for Flan-T5.} }
\end{table}

%%%%%%%%%flan t5 tables ends %%%%%%%%%%

\begin{table}[]
\centering
\caption{Performance comparison of (Others source and Others target) baseline and \textit{ConVerSum} based on mT5 on CrossSum. When we generate candidate summaries using mT5, we choose mT5 as the baseline model
to compare our model with it. LaSE, designed specifically for CLS evaluation, illustrates the
model’s robustness on CLS, which is the best choice for CLS evaluation, and BERTScore
illustrates the semantic similarity used as a supplementary metric.}
\label{tab:m5_on_crosssum_others_others}
\begin{tabular}{lllll}
\toprule
                                & \multicolumn{2}{l}{\textbf{Baseline (mT5)}} & \multicolumn{2}{l}{\textbf{ConVerSum}} \\
\multirow{-2}{*}{\textbf{\begin{tabular}[c]{@{}l@{}}Source-target\\    (others to others)\end{tabular}}} &
  \textbf{LaSE} &
  \textbf{BERTScore} &
  \textbf{LaSE} &
  \textbf{BERTScore} \\
  \midrule
Chinese\_traditional -Ukrainian & 0.5286          & 0.8321          & \textbf{0.5624 }          & 0.8305          \\

Hindi - Uzbek                   & 0.5217                    & 0.7786          & 0.4664           & \textbf{0.7804}          \\
Punjabi - Telugu                & 0.5227                    & 0.8419          & \textbf{0.5298}           & 0.8414          \\

Japanese - Burmese              & 0.5892                    & 0.8493          & \textbf{0.6500}             & \textbf{0.8613 }         \\
Marathi - Russian               & 0.5816                    & 0.8281          & 0.5769           & 0.8238          \\

Thai-Tamil                      & 0.5674                    & 0.8584          & \textbf{0.6445}           & \textbf{0.8673 }         \\
\bottomrule             
\end{tabular}
\caption*{\textit{\textbf{Key Findings: }ConVerSum shows significantly improved performance for Japanese-Burmese, and Thai-Tamil than the baseline. However, it shows lower than baseline performance for Hindi-Uzbek, and Marathi-Russian as the generated candidates fail to show diverse quality summaries in Hindi-Uzbek, and Marathi-Russian. Otherwise, ConVerSum outperforms the baseline for the rest of the pairs.} }
\end{table}

%%%%%%%%%flan t5 table starts %%%%%%%%%%
\begin{table}[]
\centering
\caption{Performance comparison of (Others source and Others target) baseline and \textit{ConVerSum} based on Flan-T5 on CrossSum. When we generate candidate summaries using Flan-T5, we choose Flan-T5 as the baseline model
to compare our model with it. LaSE, designed specifically for CLS evaluation, illustrates the
model’s robustness on CLS, which is the best choice for CLS evaluation, and BERTScore
illustrates the semantic similarity used as a supplementary metric.}
\label{tab:ft5_on_crosssum_others_others}
\begin{tabular}{lllll}
\toprule
                                  & \multicolumn{2}{l}{\textbf{Baseline (Flan-T5)}} & \multicolumn{2}{l}{\textbf{ConVerSum}} \\
\multirow{-2}{*}{\textbf{\begin{tabular}[c]{@{}l@{}}Source-target\\    (others to others)\end{tabular}}} &
  \textbf{LaSE} &
  \textbf{BERTScore} &
  \textbf{LaSE} &
  \textbf{BERTScore} \\
  \midrule
Chinese\_traditional  - Ukrainian & 0.1944                 & 0.6020                 &\textbf{ 0.2106}           & \textbf{0.6036}          \\
 
Hindi -    Uzbek                  & 0.1588                & 0.6319                 & 0.1527           & \textbf{0.6920 }         \\
Punjabi - Telugu                  & 0.2081                 & 0.6389                &\textbf{ 0.2089}           & 0.5713          \\
 
Japanese - Burmese                & 0.2277                 & 0.5264                & \textbf{0.2512}           & 0.4219          \\
Marathi - Russian                 & 0.2060                 & 0.6394                & 0.1979           & 0.6310          \\

Thai - Tamil                      & 0.2047                 & 0.5259                 & \textbf{0.2245}           & \textbf{0.5358}          \\
\bottomrule           
\end{tabular}
\caption*{\textit{\textbf{Key Findings: }ConVerSum outperforms the baseline for LaSE except for Hindi-Uzbek, and Marathi-Russian. However, it shows significantly lower performance than ConVerSum on mT5 (mentioned in Table \mbox{\ref{tab:m5_on_crosssum_others_others}}) as the generated candidates fail to show diverse quality and languages for Flan-T5.} }
\end{table}

%%%%%%%%%flan t5 table ends %%%%%%%%%%

\subsection{Second Consideration: Training \textit{ConVerSum} on Low-Resource Language Small Datasets}
\label{subsec:Training ConVerSum on Low-Resource Language Small Datasets}
This section emphasizes how we develop \textit{ConVerSum} for CLS on XLSUM~\cite{hasan-etal-2021-xl} Dataset as we mentioned in Section \ref{subsec:datasets} to observe how \textit{ConVerSum} performs on the low-resource small-scale datasets. We choose Bengali, Thai, Burmese and Tigrinya from the XLSUM dataset. As the results using mT5 outperform Flan-T5 in Section \ref{subsubsec:Performance Comparison of ConVerSum with Baselines on Cross-Lingual Dataset}, we choose only using mT5 in the case discussed in this section. Therefore, all training setups are kept the same setup mentioned in Section \ref{subsec:Training on High-Resource Large
Scale Dataset}.

\subsection{Experimental Results on Low-Resource Small Scale Dataset}
\label{subsec:Experimental Results on low-Resource small Scale Dataset}
\subsubsection{Model Performance on XLSUM Test Set}
The performance of \textit{ConVerSum} based on mT5 for training on four separate low-resource languages - Bengali, Burmese, Thai and Tigrinya on the test set of the XLSUM dataset is shown in Table \ref{tab:xlsum-test}.

\begin{table}[]
\centering
\caption{ Performance of baseline and \textit{ConVerSum} based on mT5 on the Test set of XLSUM for four separate languages-Bengali, Burmese, Thai, Tigrinya. As we generate candidate summaries using mT5,
we choose mT5 as the baseline model to compare \textit{ConVerSum} with it. LaSE, designed for specifically CLS evaluation, illustrates the
model’s robustness on CLS, which is the best choice for CLS evaluation, and BERTScore illustrates the semantic similarity used as a
supplementary metric.}
\label{tab:xlsum-test}
\begin{tabular}{lllll}
\toprule
\textbf{Training Datasets} &
  \begin{tabular}[c]{@{}l@{}}\textbf{Evaluation }\\ \textbf{Metrics}\end{tabular} &
  \textbf{Baseline(mT5)} &
  \textbf{ConVerSum} &
  \textbf{Findings} \\ \midrule
 &
  LaSE &
  0.5872 &
  \textbf{0.6228} &
   \\
\multirow{-2}{*}{XLSUM-Bengali} &
  BERTScore &
  0.8658 &
\textbf{  0.8690} &
   \\
 &
  LaSE &
  0.5579 &
  \textbf{0.5805} &
   \\
\multirow{-2}{*}{XLSUM-Burmese} &
  BERTScore &
  0.8759 &
  \textbf{0.8798} &
  \\
 &
  LaSE &
  0.4384 &
  \textbf{0.5055} &
   \\
\multirow{-2}{*}{XLSUM-Thai} &
  BERTScore &
  0.8475 &
  0.8472 &
  \\
 &
  LaSE &
  0.5265 &
  \textbf{0.5436} &
   \\
\multirow{-2}{*}{XLSUM-Tigrinya} &
  BERTScore &
  0.8747 &
  \textbf{0.8768} &
  \multirow{-8}{*}{\textit{\begin{tabular}[c]{@{}l@{}}\textit{ConVerSum} outperforms \\ baselines  for all cases\end{tabular}}}\\
  \bottomrule
\end{tabular}
\end{table}

\subsubsection{Performance Comparison of \textit{ConVerSum} Trained on XLSUM-Bengali with
Baseline on CrossSum Dataset}
\label{subsubsec:Performance Comparison of ConVerSum with Baselines on Cross-Lingual Dataset_2}

To evaluate the performance of \textit{ConVerSum} and compare it with the baselines mentioned in Section \ref{subsec:baseline}, we consider two directions - source low-resource to target another language and source any language to target low-resource language. Table \ref{tab:xlsum_bn_cross}, Table \ref{tab:xlsum_thai_cross}, Table \ref{tab:xlsum_burmese_cross}, and Table \ref{tab:xlsum_tigrinya_cross} demonstrate the performance comparison for these combinations for training \textit{ConVerSum} on Bengali, Thai, Burmese and Tigrinya, respectively.

\begin{table}[]
\centering
\caption{Performance Comparison of \textit{ConVerSum} Trained on XLSUM-Bengali with
Baseline on CrossSum Dataset}
\label{tab:xlsum_bn_cross}
\begin{tabular}{lllll}
\toprule
                         & \multicolumn{2}{l}{\textbf{Baseline (mT5})} & \multicolumn{2}{l}{\textbf{Proposed  Model}} \\
\multirow{-2}{*}{\textbf{Source-Target}} &
  \textbf{LaSE} &
  \textbf{BERTScore} &
  \textbf{LaSE} &
  \textbf{BERTScore} \\
  \midrule
\textbf{Bengali-English} & 0.5168          & 0.8263          & 0.5148           & 0.8257           \\
 
\textbf{Bengali-Arabic}  & 0.5650           & 0.8229         & 0.5535           & 0.8205           \\
\textbf{Bengali-Marathi} & 0.6197           & 0.8493          & \textbf{0.6357 }          & \textbf{0.8545}           \\
\textbf{Ukrainian-Bengali}        & 0.6110           & 0.8493          & \textbf{0.6114}           & 0.8470           \\
\textbf{Hausa-Bengali }           & 0.5831           & 0.8493          & \textbf{0.6846}           & \textbf{0.8629}        \\
\bottomrule
\end{tabular}
\caption*{\textit{\textbf{Key Findings: }ConVerSum shows significant improvement in performance for Hausa-Bengali though lower than baseline performance for Bengali-Arabic.
However, it shows almost similar to the baseline performance for Bengali-English. The generated candidates fail to show diverse quality summaries for Arabic. ConVerSum outperforms baseline for the rest of the pairs.} }
\end{table}

\begin{table}[]
\centering
\caption{Performance Comparison of \textit{ConVerSum} Trained on XLSUM-Thai with Baseline on CrossSum Dataset}
\label{tab:xlsum_thai_cross}
\begin{tabular}{lcccc}
\toprule
\multirow{2}{*}{\textbf{Source-Target}} & \multicolumn{2}{c}{\textbf{Baseline(mT5)}} & \multicolumn{2}{c}{\textbf{Proposed Model}} \\
             & \textbf{LaSE} & \textbf{BERTScore} & \textbf{LaSE} & \textbf{BERTScore }\\
\midrule
\textbf{Thai-English} & 0.5078    & 0.8298    & \textbf{0.5223}    & \textbf{0.8315}    \\
\textbf{Thai-Bengali} & 0.5937    & 0.8445    & \textbf{0.6447}    & \textbf{0.8487}    \\
\textbf{Thai-Marathi} & 0.5998    & 0.8411    & \textbf{0.6677}    & \textbf{0.8465}    \\
\textbf{Uzbek-Thai}   & 0.6672    & 0.9305    & \textbf{0.6672}    & \textbf{0.9390}    \\
\textbf{Amharic-Thai} & 0.3783    & 0.8340    & \textbf{0.4310}    & 0.8268   \\
\bottomrule
\end{tabular}
\caption*{\textit{\textbf{Key Findings: }ConVerSum outperforms baseline for all of the pairs and shows significant improvement for Thai-Bengali, Thai-Marathi, Amharic-Thai.} }
\end{table}

\begin{table}[]
\centering
\caption{Performance Comparison of \textit{ConVerSum} Trained on XLSUM-Burmese with Baseline on CrossSum Dataset}
\label{tab:xlsum_burmese_cross}
\begin{tabular}{lllll}
\toprule
\multirow{2}{*}{\textbf{Source-Target}} & \multicolumn{2}{l}{\textbf{Baseline(mT5)}} & \multicolumn{2}{l}{\textbf{Proposed Model}} \\
                & \textbf{LaSE} & \textbf{BERTScore} & \textbf{LaSE} & \textbf{BERTScore} \\
\midrule
\textbf{Burmese-English} & 0.5394    & 0.8287    & \textbf{0.5422}    & 0.8283    \\
\textbf{Burmese-Bengali} & 0.6146    & 0.8400    & 0.5973    & \textbf{0.8448}    \\
\textbf{Burmese-Marathi} & 0.4247    & 0.8371    & \textbf{0.6821}    & \textbf{0.8648}    \\
\textbf{Uzbek-Burmese}   & 0.4918   & 0.8335    & 0.4794    & \textbf{0.8431}    \\
\textbf{Amharic-Burmese} & 0.6345   & 0.8689    & 0.6054    & 0.8599 \\
\bottomrule
\end{tabular}
\caption*{\textit{\textbf{Key Findings: }ConVerSum shows significant improvement than baseline performance for Burmese-Marathi, both are low-resource languages and also outperform the Burmese-English. Moreover, our model shows better semantically similar performance than the baseline for Burmese-Bengali and Uzbek-Burmese. However, our model performs lower than the baseline only for Amharic-Burmese.} }
\end{table}

\begin{table}[]
\centering
\caption{Performance Comparison of \textit{ConVerSum} Trained on XLSUM-Tigrinya with
Baselines on CrossSum Dataset}
\label{tab:xlsum_tigrinya_cross}
\begin{tabular}{lcccc}
\toprule
                          & \multicolumn{2}{c}{\textbf{Baseline(mT5)}} & \multicolumn{2}{c}{\textbf{Proposed Model}} \\
\multirow{-2}{*}{\textbf{Source-Target}} &
  \textbf{LaSE} &
  \textbf{BERTScore} &
  \textbf{LaSE} &
  \textbf{BERTScore} \\
  \midrule
\textbf{Tigrinya-English} & 0.4478        & 0.7935       & 0.3758           & \textbf{0.7937 }         \\
 
\textbf{Tigrinya-Arabic}  & 0.4137        & 0.7837       & \textbf{0.5241 }          & \textbf{0.7966}          \\
\textbf{Tigrinya-Marathi} & 0.4185        & 0.7762       & \textbf{0.6423}           & \textbf{0.7916 }    \\
\textbf{Chinese\_simplified-Tigrinya} & 0.4390 & 0.8288 & \textbf{0.5497} & \textbf{0.8288} \\
\textbf{Somali-Tigrinya} & 0.2241 & 0.7829 & \textbf{0.5537} & 0.7801 \\
\bottomrule
\end{tabular}
\caption*{\textit{\textbf{Key Findings: }ConVerSum shows significant improvement over baseline performance for all pairs except Tigrinya-English. } }
\end{table}

\subsection{Performance Comparison of \textit{ConVerSum} and LLMs for Low Resource Languages by Prompting}

Evaluating the performance of the state-of-art LLMs on various datasets and numerous NLP tasks is a trending research~\cite{laskar-etal-2023-systematic, benllm}. Inspired by this trend, in this section, we compare the performance of \textit{ConVerSum} to cutting-edge LLMs - ChatGPT-3.5, ChatGPT-4o, and Gemini 1.5 Pro to measure \textit{ConVerSum}'s efficiency in generating high-quality summaries in a variety of low-resource languages. To assess LLMs' proficiency, we prompt them to rate their level of confidence between 1 to 10 in generating accurate cross-lingual summaries in 45 different languages. Responses are repeatedly generated to ensure clarity. All LLMs rate their capability as low \textit{(1-4)} for CLS on low-resource languages. We choose three language pairs mentioned in Table \ref{tab:comparison_with_llm} based on their ratings for this experiment.    

\begin{tcolorbox}[colframe=black, colback=white, left=1em, right=1em, top=1em, bottom=1em]
\textit{"How confident are you to generate high-quality cross-lingual summary concisely and informatively for low-resource languages? Find the list of languages - Amharic, Arabic, Azerbaijani, Bengali, Burmese, Chinese\_simplified, Chinese\_traditional, English, French, Gujarati, Hausa, Hindi, Igbo, Indonesian, Japanese, Kirundi, Korean, Kyrgyz, Marathi, Nepali, Oromo, Pashto, Persian, Pidgin, Portuguese, Punjabi, Russian, Scottish\_gaelic, Serbian\_cyrillic, Serbian\_latin, Sinhala, Somali, Spanish, Swahili, Tamil, Telugu, Thai, Tigrinya, Turkish, Ukrainian, Urdu, Uzbek, Vietnamese, Welsh, Yoruba. Rate your confidence level for cross-lingual summarization on a scale of 1 to 10 for the given languages."} 
\end{tcolorbox}

Then, we use manual prompting for LLMs which is free of cost. For our selected language pairs, we provide the following prompt where the source document is provided from CrossSum~\cite{bhattacharjee-etal-2023-crosssum} test set and extract the responses manually from LLMs to measure the LaSE and BERTScore and compare those with \textit{ConVerSum} which are mentioned in Table \ref{tab:comparison_with_llm}.

\begin{tcolorbox}[colframe=black, colback=white, left=1em, right=1em, top=1em, bottom=1em]
\textit{"Summarize the given text in \textbf{target language}, preferably in 80 words, concisely and informative. .. (source document is given here)" }
\end{tcolorbox}

\begin{table}[]
\centering
\caption{Some Performance Measures on LLM vs \textit{ConVerSum} for Low-resource Languages}
\label{tab:comparison_with_llm}
\begin{tabular}{lllllllll}
\toprule
\multirow{2}{*}{\textbf{Source-Target}} &
  \multicolumn{2}{c}{\textbf{ChatGPT 3.5}} &
  \multicolumn{3}{c}{\textbf{ChatGPT 4o}} &
  \multicolumn{2}{c}{\textbf{ConVerSum}} &
  \multicolumn{1}{c}{\multirow{2}{*}{\textbf{Gemini 1.5 Pro}}} \\
 &
  \textbf{LaSE} &
  \textbf{BERTScore} &
  \textbf{LaSE} &
  \multicolumn{2}{l}{\textbf{BERTScore}} &
  \textbf{LaSE} &
  \textbf{BERTScore} &
  \multicolumn{1}{c}{} \\
  \midrule
\textbf{Amharic-Igbo} &
  \multicolumn{2}{l}{\begin{tabular}[c]{@{}l@{}}Attempts to \\ generate \\ monolingual\\  summary\end{tabular}} &
  0.7264 &
  \multicolumn{2}{l}{0.7364} &
  0.6377 &
  0.7836 &
  \multirow{3}{*}{\begin{tabular}[c]{@{}l@{}}Unable \\ to summarize\end{tabular}} \\
\textbf{Tigrinya-Arabic} &
  0.2798 &
  0.7481 &
  0.5246 &
  \multicolumn{2}{l}{0.7592} &
  0.5241 &
  0.7936 &
   \\
\textbf{Burmese-Bengali} &
  \multicolumn{2}{l}{\begin{tabular}[c]{@{}l@{}}Attempts to \\ generate\\  monolingual \\ summary.\\  After a few \\ prompts, \\ generate\\  inaccurate CLS\end{tabular}} &
  \multicolumn{3}{l}{\begin{tabular}[c]{@{}l@{}}Input a document \\ of 699 words and\\ unable to handle \\ the long text\end{tabular}} &
  0.6030 &
  0.8452 & \\
  \bottomrule
\end{tabular}
\end{table}

\subsubsection{Limitations of Prompt-based Approach}
Although this manual prompting and extracting approach is budget-friendly, it is very time-consuming for broad-level experiments. Moreover, this approach limits handling long documents. Besides, it has confined us to conducting thorough experiments. For this reason, we experiment with using API to conduct a thorough experiment discussed in Section \ref{sec:api_gpt}. We only choose GPT 4o latest version model as it is a more powerful LLM than GPT 3.5 or Gemini.

\subsection{Performance Comparison of GPT-4o using API and \textit{ConVerSum}}
\label{sec:api_gpt}
This section offers a comparison of \textit{ConVerSum} with GPT-4o for a range of language pairings in a zero-shot setup. Zero-shot learning refers to the model's capacity to provide summaries for language pairings that were not explicitly included in its training data. For this purpose, we use the GPT-4o model version gpt-4o-2024-05-13\footnote{\url{https://platform.openai.com/docs/models/gpt-4o}}, which is the most recent version and represents the model's status as of October 2023. This version can handle up to 128,000 tokens and a maximum context window of 4,096 tokens. Furthermore, we want to explore how GPT 4o performs on a one-shot setup which is a form of machine learning in which a model learns to recognize objects or execute tasks based on only one training sample per class.   

% \usepackage{multirow}
\begin{table}[]
\centering
\caption{Performance comparison of \textit{ConVerSum} and GPT 4o in zero-shot setup using API}
\label{tab:gpt4o api}

\begin{tabular}{cccccc} 
\toprule
 & & 
  \multicolumn{2}{c}{\textbf{GPT 4o (current version)}} &
  \multicolumn{2}{c}{\textbf{ConVerSum}} \\ 
\multirow{-2}{*}{\textbf{Source-Target}} &
  \multirow{-2}{*}{\begin{tabular}[c]{@{}c@{}}\textbf{No of} \\ \textbf{Samples}\end{tabular}} &
  \textbf{LaSE} &
  \textbf{BERTScore} &
  \textbf{LaSE} &
  \textbf{BERTScore} \\
  \midrule
 
Burmese-Bengali                 & 14          & 0.7272 & 0.7995 & 0.6030 & \textbf{0.8452 }\\

Amharic-Igbo                    & 3           & 0.7142 & 0.7370 & 0.6377 & \textbf{0.7836} \\

Tigrinya-Arabic                 & 2           & 0.5903 & 0.7538 & 0.5241 &\textbf{ 0.7936} \\

\textbf{Japanese-Burmese}                & 14 & \textbf{0.5620 }& \textbf{0.7853} & \textbf{0.6500} & \textbf{0.8613} \\
 
Thai-Tamil                      & 40          & 0.6784 & 0.7943 & 0.6445 & \textbf{0.8673} \\
 
\textbf{Hausa-Bengali}                   & 23 & \textbf{0.6532} & \textbf{0.7759} & \textbf{0.6846} & \textbf{0.8629} \\

\textbf{Chinese\_simplified-Tigrinya}    & 1           & \textbf{0.3178} & \textbf{0.7151} & \textbf{0.5497} &\textbf{ 0.8288} \\
 
\textbf{Burmese-Marathi}                 & 9           & \textbf{0.6606} & \textbf{0.7912} & \textbf{0.6821} & \textbf{0.8648} \\
 
Thai-Marathi                    & 20          & 0.6743 & 0.7937 & 0.6677 & \textbf{0.8465} \\

Chinese\_simplified-English     & 486         & 0.8156 & 0.8162 & 0.5361 & \textbf{0.8437} \\
 
Chinese\_traditional-Ukrainian & 143         & 0.6520 & 0.8022 & 0.5624 & \textbf{0.8305} \\

\textbf{Uzbek-Thai}                      & 3           & \textbf{0.5857} & \textbf{0.7173} & \textbf{0.6672} & \textbf{0.9390}\\
\bottomrule
\end{tabular}
\end{table}

\subsubsection{Performance Analysis on Zero-Shot Setup}
Table \ref{tab:gpt4o api} illustrates the performance comparison between GPT-4o and \textit{ConVerSum} across various language pairs. We conduct experiments on 12 separate language pairs. We consider several combinations for choosing language pairs as per GPT 4o’s confidence level such as low-low, high-low, high-high, mid-mid, and so on.

Among 12 language pairs, \textit{ConVerSum} 
outperforms 5 language pairs GPT 4o in which
3 pairs are not included in the list of the high confidence level of CLS for GPT-4o.  and 2 pairs have at least one between each source and target pair from a high confidence level for GPT-4o. 
\textit{ConVerSum} shows almost similar performance for  1 pair where both source and target are from mid-confident languages by GPT 4o.

\textit{ConVerSum} shows a clear benefit in dealing with low-resource languages, as observed in pairs like Burmese-Marathi and Uzbek-Thai, Hausa-Bengali which pairs are not included in list of the high confidence level of CLS for GPT-4o. 

There are also two language pairs in which \textit{ConVerSum} outperforms GPT 4o - Japanese-Burmese and Chinese\_simplified-Tigrinya. Among these languages, we find at least one between each source and target pairs such as Japanese and Chinese\_simplified from the high confidence languages for GPT 4o.

This is seen by language pairs like Chinese\_simplified-English, where it received the highest LaSE score of 0.8156, indicating stronger generalization for high-resource languages.
\textit{ConVerSum} performs significantly lower than GPT 4o for source high-confident and target high-confident language pair combinations.

The proposed model's constant outperformance in BERTScore, despite variations in LaSE scores, indicates its power in capturing semantic and contextual complexities, especially in language pairings with limited resources where GPT-4o's performance is somewhat weaker. From this discussion, we conclude that in situations when data availability is constrained, how well the model can perform CLS tasks.

\subsubsection{Performance Analysis on One-Shot Setup}
\label{subsec:oneshot}
We want to explore the performance of the GPT 4o model in a one-shot setup. For this purpose, we choose 3 low-resource language pairs. We provide only one example from the train set of the CrossSum dataset from the respective language-pair datasets. Table \ref{tab:one shot} illustrates the performance comparison among GPT 4o's zero-shot, one-shot and \textit{ConVerSum}.

% \usepackage{multirow}
\begin{table}[]
\centering
\caption{Performance comparison of \textit{ConVerSum} and GPT 4o in zero-shot and one-shot setup using API}
\label{tab:one shot}
\begin{tabular}{cccccccc}
\toprule
                &    & \multicolumn{4}{c}{\textbf{GPT 4o (current model)}} & \multicolumn{2}{c}{} \\
 &
   &
  \multicolumn{2}{c}{\textbf{Zero Shot}} &
  \multicolumn{2}{c}{\textbf{One shot}} &
  \multicolumn{2}{c}{\multirow{-2}{*}{\textbf{ConVerSum}}} \\
\multirow{-3}{*}{\textbf{Source-Target}} &
  \multirow{-3}{*}{\begin{tabular}[c]{@{}c@{}}\textbf{No of} \\ \textbf{Samples}\end{tabular}} &
  \textbf{LaSE} &
  \textbf{BERTScore} &
  \textbf{LaSE }&
  \textbf{BERTScore} &
 \textbf{ LaSE} &
  \textbf{BERTScore} \\
  \midrule

Burmese-Bengali & 14 & 0.7272    & 0.7995   & 0.7204   & 0.8001   & 0.6030    & \textbf{0.8452 }  \\
Amharic-Igbo    & 3  & 0.7142    & 0.7370   & 0.6871   & 0.7352   & 0.6377    & \textbf{0.7836}   \\

\textbf{Tigrinya-Arabic} & 2  & 0.5903    & 0.7538   & 0.\textbf{5222}   & \textbf{0.7551 }  & \textbf{0.5241}    & \textbf{0.7936}  \\
\bottomrule
\end{tabular}
\end{table}

From Table \ref{tab:one shot}, we observe that the one-shot performance of GPT 4o for LaSE has been decreased than zero-shot for GPT 4o. As CLS can be challenging to LLMs when there’s not enough contextual understanding or fine-tuning for the specific language pairs, they struggle to perform well when it involves nuances specific to certain languages like idioms or syntax.
For 1 language pair (Tigrinya-Arabic), \textit{ConVerSum} outperforms GPT 4o in one-shot setup.
\textit{ConVerSum} outperforms BERTScore for all pairs we experimented with highlighting capturing semantic and contextual complexities. 

\subsection{Discussion on LLM for CLS}
Based on our experiments on LLM for low-resource languages, we have some observations on LLMs. These are as follows:

\subsubsection{Gemini 1.5 Pro: } Gemini only supports a few languages and frequently fails to correctly detect the specified source language. It struggles to produce summaries from documents in low-resource source languages and is equally unable to provide summaries in low-resource target languages, demonstrating its shortcomings in multilingual and low-resource language contexts.

\subsubsection{ChatGPT 3.5: } When using ChatGPT 3.5 to handle low-resource languages, the model is frequently unable to provide high-quality, accurate summaries. For several low-resource languages, it generates monolingual summaries rather than CLS. Furthermore, the model produces inefficient summaries even after several prompts, showing its difficulties in properly digesting and summarizing content in low-resource language contexts.

\subsubsection{ChatGPT 4o by Prompting Approach: } When reviewing the capabilities of the latest and most powerful LLM, ChatGPT-4o, which was launched on May 2024, it is stated that, while this model is capable of handling low-resource languages, it still has limits. It cannot handle large documents (more than 700 words) for several low-resource languages, such as Burmese, emphasizing areas where further improvements are required.

\subsubsection{ChatGPT 4o Using API:} GPT 4o outperforms for several language pairs specifically it shows significantly better performance for high-resource languages. However, GPT-4o’s performance is somewhat weaker for language pairings with limited resources where it involves idioms or syntax of those languages.  

\section{Result Analysis in Different Aspects}
This section analyzes our findings based on different aspects like low and high-resource scenarios, and large and small-scale dataset scenarios. Furthermore, the detailed analysis based on the evaluation metrics is discussed. 

\subsection{For Training on High-Resource Large-Scale Dataset}
\subsubsection{Comparison between mT5 and Flan-T5 based Model}
For all cases, the Flan T5 model as a baseline and the proposed model using Flan T5 show significantly lower performance than \textit{ConVerSum} using mT5 and also for the mT5 itself as a baseline. The robustness of \textit{ConVerSum} heavily relies upon the quality of candidate generation used in contrastive learning.
As there is no Flan-T5 checkpoint specifically fine-tuned on cross-lingual pairs, the generated candidates fail to show diverse quality and languages for Flan-T5. Though Flan-T5 is multilingual, it sometimes shows biases in high-resource languages like English.

\subsubsection{\textit{ConVerSum} using mT5}
While training on the high-resource large-scale dataset as discussed in Section \ref{subsec:Training on High-Resource Large
Scale Dataset}, \textit{ConVerSum} outperforms the baseline in LaSE for most source-target language pairs, except Bengali-English and Tigrinya-English, where it performs similarly to the baseline. For BERTScore, our model's performance is nearly identical to the baseline mT5 for a few cases and others outperform the baseline.

\subsection{For Training on Low-Resource Small-Scale Dataset}
While training on the low-resource small-scale dataset as discussed in Section \ref{subsec:Training ConVerSum on Low-Resource Language Small Datasets}, \textit{ConVerSum} outperforms the baseline for most of the language pairs, and only for a few instances, it shows almost identical performance to baseline.

\subsection{Comparison between LaSE and BERTScore for CLS}
In almost all cases, \textit{ConVerSum} outperforms the baseline for LaSE rather than BERTScore except in a few cases. For CLS evaluation, LaSE is generally considered a better choice than BERTScore as LaSE prioritizes meaning and fluency across languages, which is more important than just word-level similarity.
BERTScore only focuses on higher semantic similarity. There are some results where BERTScore improves and LaSE doesn’t. This scenario indicates the meaning similarity is improved but is unable to guarantee the summary length and target language. For example, we want to generate a summary for an English source document to target Bengali. We get two summaries as mentioned in Table \ref{tab:lase_vs_bs}. 

\begin{table}[]
\centering
\caption{Comparison between LaSE and BERTScore for CLS}
\label{tab:lase_vs_bs}
\begin{tabular}{ccc}
\toprule
\textbf{No} &
  \textbf{Summary} &
  \textbf{Analysis} \\ \midrule
1 &
  \begin{tabular}[c]{@{}c@{}}\includegraphics[width=0.4\linewidth]{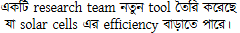}\end{tabular} &
  \begin{tabular}[c]{@{}c@{}}Word overlap results in a high BERTScore, \\ although Bengali and English are mixed together.\\ Lower LaSE for language confidence.\end{tabular} \\
2 &
  \begin{tabular}[c]{@{}c@{}}\includegraphics[width=0.4\linewidth]{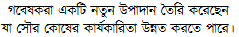}\end{tabular} &
  \begin{tabular}[c]{@{}c@{}}Lower BERTScore due to lack of direct\\ word overlap but provides a correct \\ CLS higher LaSE
  \end{tabular}
  \\ \bottomrule
\end{tabular}
\end{table}

From Table \ref{tab:lase_vs_bs}, Summary 1 still receives a high BERTScore though there is a mixing of source and target languages. The LaSE score is lower in this case because it ensures summaries are generated in the target language.
Summary 2 receives a lower BERTScore as there is a lack of direct word overlap though it provides a correct cross-lingual summary. Therefore, LaSE receives a high score in this case maintaining its criteria for CLS evaluation~\cite{bhattacharjee-etal-2023-crosssum}. This dual-metric evaluation delivers essential insights to satisfy the demands of different application scenarios.

\subsection{Overall Discussion}
\textit{ConVerSum} performs better in both high- and low-resource scenarios, and large- and small-scale datasets, and consistently beats the baseline models. It even performs better than GPT-3.5 and almost as well as GPT-4o in low-resource language settings, outperforming the capabilities of current LLMs. The observation that our model performs well even with limited training data for low-resource languages highlights the efficacy of the contrastive learning approach used. These outcomes validate the robustness and flexibility of our model, highlighting the ability of contrastive learning to improve model performance even in resource-constrained scenarios.

\section{Conclusion}
The paper introduces \textit{ConVerSum}, a novel contrastive learning-based CLS approach. By extracting and generalizing semantic representations across languages, this technique enhances CLS's adaptability and durability. A comprehensive comparison with state-of-the-art models and various cross-lingual language pairs proves the effectiveness of the approach. This strategy works especially well when there is no CLS parallel data available, as traditional approaches might not be able to produce accurate findings. \textit{ConVerSum} improves cross-lingual summaries by utilizing the unique characteristics of contrastive learning, therefore establishing a new standard for further studies. Nonetheless, there are several difficulties, such as the large computational capacity needed for training and the requirement for a large number of candidate summaries from various language sets. In the future, there are potential improvements in performance, adjustments to methods, and the use of guidelines to guarantee fairness and biases.

\begin{acks}
This work was funded by the Research and Innovation Centre for Science and Engineering (RISE), BUET.
\end{acks}

\bibliographystyle{ACM-Reference-Format}
\bibliography{sample-base}

%\appendix

\end{document}